\documentclass[letterpaper, 10 pt, journal, twoside]{IEEEtran}

\usepackage{amsmath,amsfonts, amssymb}
\usepackage{hyperref}
\usepackage[linesnumbered,ruled,vlined]{algorithm2e} 
\usepackage{array}
\usepackage{textcomp}
\usepackage{stfloats}
\usepackage{url}
\usepackage{verbatim}
\usepackage{graphicx}
\usepackage{cite}
\usepackage{subcaption} 
\usepackage{adjustbox}
\usepackage{booktabs}
\usepackage{threeparttable}
\usepackage{tabularx}
\usepackage{xcolor}
\usepackage{soul}
\usepackage[letterpaper, top=60pt, bottom=48pt,  left=48pt, right=48pt]{geometry}

\newcommand{\bs}[1]{\boldsymbol{#1}}

\ifCLASSINFOpdf
\else
\fi
\begin{document}
%
\title{SGBA: Semantic Gaussian Mixture Model-Based LiDAR Bundle Adjustment}
%
%
%

\author{Xingyu Ji, Shenghai Yuan*, \IEEEmembership{Member,~IEEE}, Jianping Li, \IEEEmembership{Member,~IEEE}, \\Pengyu Yin, Haozhi Cao, Lihua Xie, \IEEEmembership{Fellow,~IEEE} 
\thanks{Manuscript received: June, 23, 2024; Revised September, 10, 2024; Accepted October, 02, 2024.}
\thanks{This paper was recommended for publication by Editor Javier Civera upon evaluation of the Associate Editor and Reviewers' comments.}
\thanks{This work is supported by National Research Foundation, Singapore, under its Medium-Sized Center for Advanced Robotics Technology
Innovation.} 
\thanks{* Corresponding author. All authors are with the Centre for Advanced Robotics Technology Innovation (CARTIN), School of Electrical and Electronic Engineering, Nanyang
Technological University, Singapore. Email: {\tt\small \{xingyu001, shyuan, jianping.li, pengyu001, haozhi.cao, elhxie\}@ntu.edu.sg}}%
\thanks{Digital Object Identifier (DOI): see top of this page.}
}

%
%

\markboth{IEEE Robotics and Automation Letters. Preprint Version. Accepted October, 2024}
{Ji \MakeLowercase{\textit{et al.}}: SGBA: Semantic Gaussian Mixture Model-Based LiDAR Bundle Adjustment}

%



\maketitle


\IEEEpubid{\begin{minipage}{\textwidth}\ \\[55pt]
\centering
\fbox{%
\parbox{\dimexpr\textwidth-2\fboxsep-2\fboxrule}{%
\centering
\footnotesize
This work has been accepted for publication in IEEE Robotics and Automation Letters (RAL) © 2025 IEEE.\
Personal use of this material is permitted. However, permission must be obtained from IEEE for all other uses,\
including reprinting or redistribution, creating derivative works, or reuse of any copyrighted components of this work in other media.
}%
}
\end{minipage}}

\begin{abstract}
LiDAR bundle adjustment (BA) is an effective approach to reduce the drifts in pose estimation from the front-end. Existing works on LiDAR BA usually rely on predefined geometric features for landmark representation. This reliance restricts generalizability, as the system will inevitably deteriorate in environments where these specific features are absent. To address this issue, we propose SGBA, a LiDAR BA scheme that models the environment as a semantic Gaussian mixture model (GMM) without predefined feature types. This approach encodes both geometric and semantic information, offering a comprehensive and general representation adaptable to various environments. Additionally, to limit computational complexity while ensuring generalizability, we propose an adaptive semantic selection framework that selects the most informative semantic clusters for optimization by evaluating the condition number of the cost function. Lastly, we introduce a probabilistic feature association scheme that considers the entire probability density of assignments, which can manage uncertainties in measurement and initial pose estimation. We have conducted various experiments and the results demonstrate that SGBA can achieve accurate and robust pose refinement even in challenging scenarios with low-quality initial pose estimation and limited geometric features. We plan to open source the work for the benefit of the community \url{https://github.com/Ji1Xingyu/SGBA}.
\end{abstract}


%
\IEEEpeerreviewmaketitle

\section{Introduction}

LiDAR odometry has been extensively studied for its crucial role in providing pose information for field robotics. The typical approach involves pairwise registration to align the current LiDAR scan with the local map for pose estimation. However, due to the inherent limitations of pairwise registration, LiDAR odometry inevitably accumulates drifts over long durations \cite{yuan2021survey}. To correct these drifts, researchers proposed optimizing multiple robot poses and landmark states simultaneously, referred to as bundle adjustment (BA). BA consists of three essential parts: landmark representation, feature association, and error minimization \cite{triggs2000bundle}.

\begin{figure}[!t]
\centering
\includegraphics[width=3.4in]{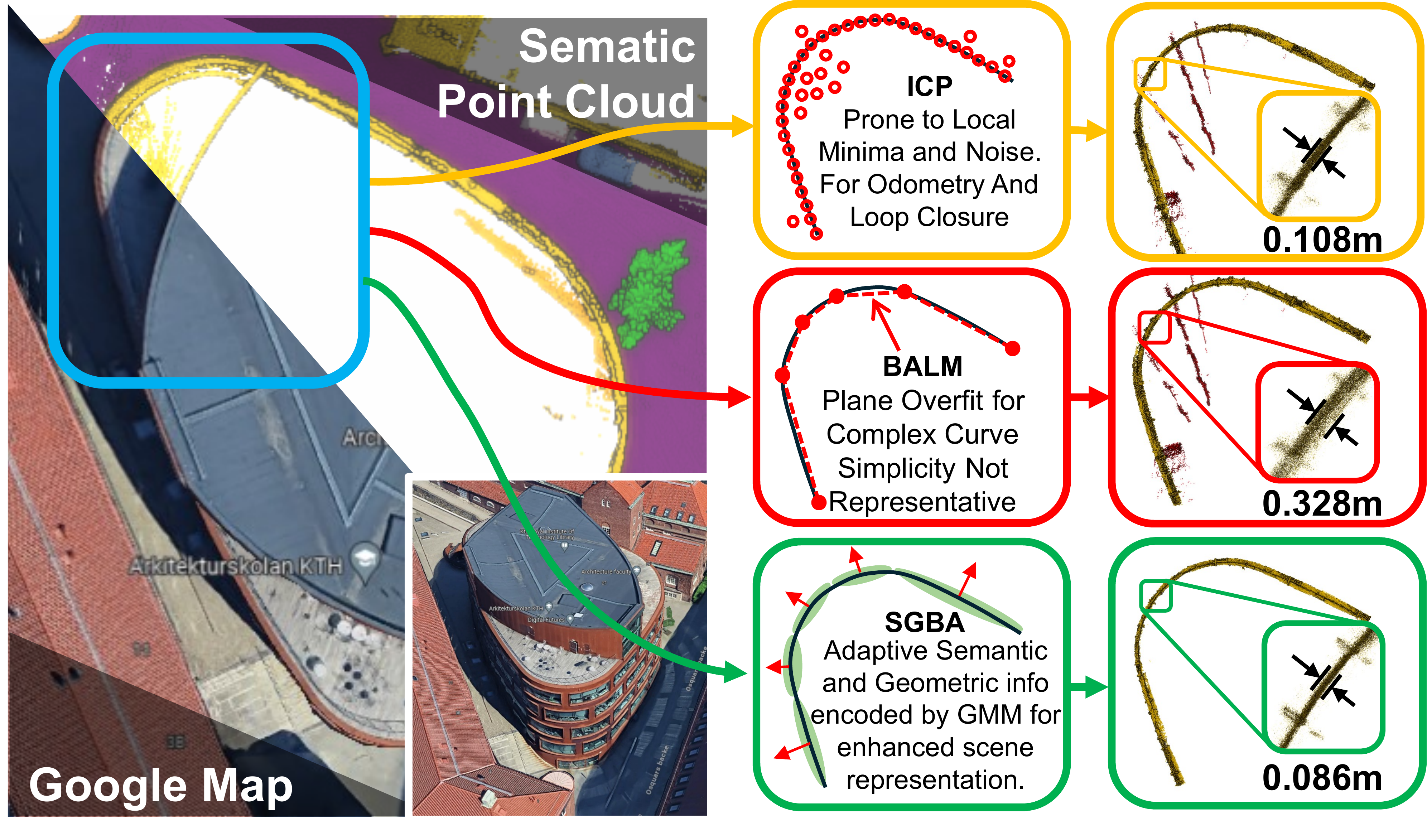}
\vspace{-5pt}
\caption{Comparison of different landmark representation methods and their BA performance.  }
\label{fig_performance_comp}
\vspace{-15pt}
\end{figure}

In visual BA, landmarks are 3D points projected from image feature points, with feature association achieved through point correspondences by descriptors like SIFT \cite{triggs2000bundle}. However, in LiDAR BA, selecting 3D points as landmarks is impractical due to the sparse and non-repetitive nature of LiDAR data \cite{balm}. Most existing works on LiDAR BA extract predefined geometric features as landmarks, such as planar and edge features \cite{balm2, plc, ef, multiview}. This approach lacks generalizability in environments without these specific features \cite{pss_ba}, as shown in Fig. \ref{fig_performance_comp}. Poor initial pose estimates can bias these geometric features, leading to system failure. For feature association, current LiDAR BA methods adopt deterministic feature association, where measurements are hard-assigned to nearby landmarks \cite{balm2, ef, plc, multiview, pss_ba}. These methods typically rely on ICP variants for optimization, which cannot guarantee reaching a global optimum. Poor initial estimates can cause early convergence to local optima \cite{sem_asso}. 

In conclusion, the main challenges for LiDAR BA are effective landmark representation in diverse environments and accurate feature association to avoid early convergence to local optima. 

Incorporating semantic information is a viable solution to these challenges. Advances in semantic segmentation algorithms now allow efficient and accurate extraction of point-wise semantic information from LiDAR point clouds \cite{chen2019suma++, spvnas, cao2023multi}. This integration enables autonomous systems to distinguish between different types of objects, such as vehicles, pedestrians, and static infrastructure, providing a richer contextual understanding of environments. Introducing semantic constraints for feature association ensures consistency both geometrically and semantically \cite{chen2019suma++}. However, selecting all semantic layers is computationally expensive, while using specific layers limits generalizability. Thus, developing an adaptive semantic layer selection framework is essential.
In this paper, we propose a Semantic Gaussian Mixture Model-based LiDAR Bundle Adjustment (SGBA), encoding both geometric and semantic attributes. We model the environment as a semantic GMM for flexible, general landmark representation. An adaptive semantic selection framework, based on the condition number of the optimization problem, selects the most informative semantic layers, balancing efficiency and generalizability. Additionally, a soft-assign scheme for feature association considers the entire probability density of assignments, handling uncertainties and ambiguities in LiDAR data and initial pose estimates. Our contributions are summarized as follows:
\begin{itemize}
    \item \textbf{Comprehensive Landmark Representation}: Our approach models the environment as a semantic GMM, integrating geometric and semantic information into a comprehensive representation. This enhances adaptability, robustness, and generalizability across different environments without relying on predefined feature types. 
    
    \item \textbf{Adaptive Semantic Selection Framework}: We introduce a condition number-based method to adaptively select the most informative semantic layers for the BA process. This avoids predefined layers, ensuring generalizability and enhancing computational efficiency by minimizing constraints.
    \item \textbf{Probabilistic Feature Association}: We propose a soft-assign scheme that considers the entire probability density of feature assignments, effectively handling uncertainties and ambiguities in LiDAR data and initial pose estimates for more reliable pose estimation.
\end{itemize}

\section{Related Works}

In terms of landmark representation, most existing works on LiDAR BA extract predefined geometric features as landmarks. These geometric features ensure co-visibility across multiple scans and provide reliable point-landmark correspondences. Early works parameterized planar features and treated the point-to-plane distance as the cost function \cite{early}. To improve the efficiency, a novel approach was proposed which reformulated the the cost function such that its complexity is independent of the number of points \cite{ef}. This idea was extended in \cite{balm} and a closed-form derivation of the Hessian matrix for the cost function was provided for acceleration. The limitations of these works \cite{ef, balm} were discussed in \cite{multiview} and a new cost function was introduced to address these issues. \cite{balm2} introduced the concept of point cluster, drastically speeding up the BA process. The feature types were extended to line, planar, and cylinder in \cite{plc}, which is more generalized compared to planar features only. However, this method still requires predefined feature types and lacks general applicability. To address the limit of predefined geometric features, \cite{pss_ba} models the environment with polynomial surface kernels, providing a more general environment representation. 

So far, few works on LiDAR BA have integrated semantic information. Incorporating semantic information enables the system to differentiate between various object types, thus offering a more comprehensive contextual understanding of the environment. For LiDAR odometry, \cite{sloam, semantic_icp} improved the feature association with semantic constraints. In terms of semantic mapping, \cite{chen2019suma++} further proposed to filter out dynamic objects by checking semantic consistency to construct a reliable map. For global localization, \cite{yin2023outram} combined semantic information with a triangulated scene graph for more informative correspondence extraction. For LiDAR BA, we propose to model the landmarks as semantic GMM, similar to the approach in \cite{jrmpc}, which adopted a GMM representation for multiview registration. The major challenge of adopting GMM in LiDAR BA is its low efficiency in formulating and solving the cost function. We address this by incorporating semantic information in two ways: firstly, we adaptively select semantic layers for constructing the cost function to limit its dimension; secondly, we use semantic constraints to limit the complexity of feature associations. 

In terms of feature association, existing LiDAR BA methods adopt a deterministic association strategy. \cite{plc} directly assign the point with its nearest landmarks using the initial pose estimation. \cite{balm, multiview} voxelized the initial map and performed feature association within each voxel. The drawback of the deterministic association is that incorrect associations can move the optimal solution far away from the global optimum \cite{bowman2017probabilistic}. To address this issue, we propose to consider the entire probability density of the assignment, treating it as a latent variable. This probabilistic approach, similar to expectation maximization (EM) \cite{tabib2018manifold, em_icp}, can better handle uncertainties and ambiguities in the data. The major difference between our approach and existing methods is the introduction of semantic constraints, such that the correspondences are matched, ensuring both geometric and semantic consistency. 

\section{Problem Formulation}
Consider $K$ LiDAR scans $\boldsymbol{Z} = \left\{ \boldsymbol{Z}_k \right\}_{k=1}^K$ captured by a robot. Each scan $\boldsymbol{Z}_k$ consists of $N_k$ points as $\boldsymbol{Z}_k = \left \{ \hat{\boldsymbol{z}}_{ki}
\right\}_{i=1}^{N_k}$. Specifically,
$\hat{\boldsymbol{z}}_{ki} = \begin{bmatrix}
        \boldsymbol{z}_{ki}^T & s_{ki}
    \end{bmatrix}^T$
represents that each point stores its spatial coordinate $\boldsymbol{z}_{ki}$ and the semantic label $s_{ki}\in\mathbb{R}$. The corresponding poses for these scans are $\boldsymbol{T} = \left\{ \boldsymbol{T}_k \right\}_{k=1}^K$, where $\boldsymbol{T}_k \in \mathrm{SE}(3)$ is a transformation matrix composed of the rotation matrix $\bs{R}_k \in\mathrm{SO}(3)$ and position $\bs{p}_k\in \mathbb{R}^3$. Additionally, we will implicitly convert the point $\hat{\bs{z}}_{ki}$ to its homogeneous space for convenience when expressing its transformation to the global frame, denoted by $\bs{T}_k \bs{z}_{ki}$.

Suppose $J$ landmarks $\boldsymbol{L} = \left\{ \boldsymbol{l}_j \right\}_{j = 1}^J$ are visible for all the scans, the LiDAR BA problem can be interpreted as estimating the optimal $\hat{\boldsymbol{T}}$, $\hat{\boldsymbol{L}}$ based on the measurements $\boldsymbol{Z}$. This refers to the following Maximum Likelihood Estimation (MLE) problem:
\begin{equation} \label{eq_mle_ori}
\begin{split}
    \hat{\boldsymbol{T}}, \hat{\bs{L}} &= \operatorname*{argmax}_{\bs{T}, \bs{L}, \bs{Y}} \mathcal{L}\left( \bs{T}, \bs{L} \left \vert \bs{Z} \right.\right) \\
    &= \operatorname*{argmax}_{\bs{T}, \bs{L} }\log p\left( \bs{Z} ; \bs{T}, \bs{L} \right).
\end{split}
\end{equation}
However, direct optimization of this MLE problem is intractable due to the unobservable variable: the feature association relationship. To clarify, we denote this relationship as $\bs{Y} = \left\{ y_{ki}: k\in\left[1, K\right], i\in\left[1, N\right] \right\}$, where $y_{ki}=j$ implies that the measurement $\hat{\bs{z}}_{ki}$ is assigned to the landmark $\bs{l}_j$.

\section{Method}

\begin{figure*}[ht]
\centering
\includegraphics[width=6.8in]{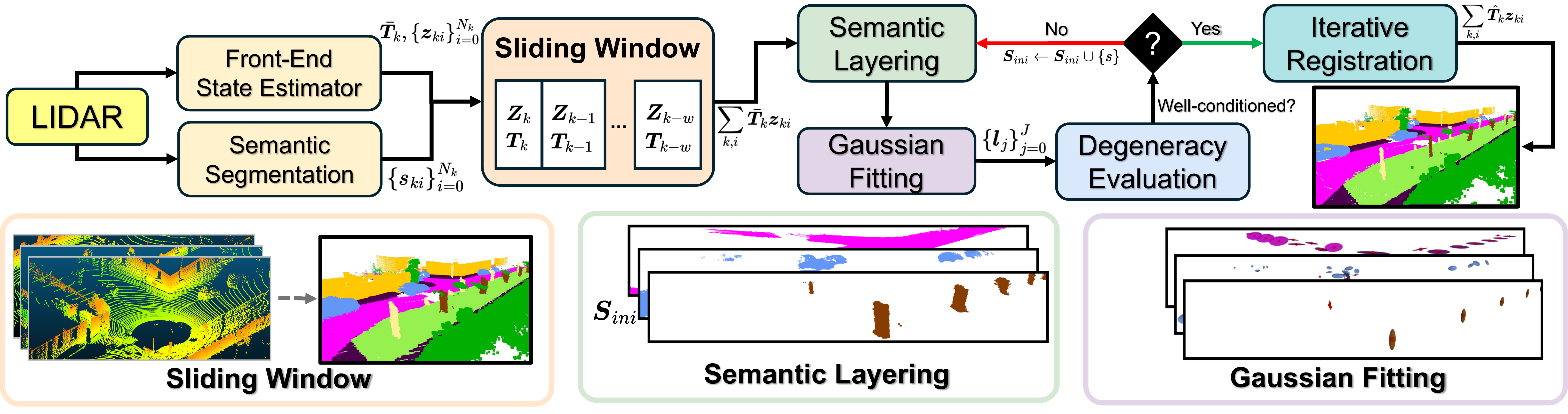}
\vspace{-5pt}
\caption{Workflow of SGBA. The system first takes the aggregated cloud as input and layers the cloud based on semantic labels. Afterward, initial Gaussians are fitted separately on each semantic layer. An initial optimization problem is constructed and evaluated for degeneracy. If degeneracy is detected, additional semantic layers are included. Finally, the optimal robot poses are estimated by iteratively solving the MLE problem \eqref{eq_mle_ori}. }
\label{fig_workflow}
\vspace{-10pt}  
\end{figure*}

\subsection{Gaussian Mixture Representation with Semantic Cues}

We propose a general model for landmark representation, denoted as Gaussian mixture representation with semantic cues (see Fig. \ref{fig_workflow}). This approach directly models the entire spatial likelihood of the environment as a semantic GMM without the need to extract predefined features. As a result, the landmark can be represented as a Gaussian under global frame: $\boldsymbol{l}_j = \mathcal{N}\left( \bs{\mu}_j, \bs{\Sigma}_j \right)$, each associated with a semantic label $s_j$. The integration of semantic cues enables the extraction of stable objects (e.g., buildings, poles) for robust and efficient pose estimation while filtering out unstable elements. Furthermore, we constrain feature association within specific semantic layers and thus can reject outliers by dismissing associations from different labels (illustrated in Fig. \ref{fig_workflow}). The criteria for selecting semantic labels are adaptively determined and will be detailed in Sec \ref{sec_label_select}.

With semantic GMM, the likelihood $p$ of a measurement $\hat{\bs{z}}_{ki}$ given its assignment to the landmark $j$ is:
\begin{equation} \label{eq_likelihood_single_measurement}
    p\left( {\bs{z}}_{ki} \left\vert \bs{T}, \bs{L}, y_{ki}=j \right. \right) = \mathbf{1}_{s_{ki} = s_j}\mathcal{N}\left( \bs{T}_k\bs{z}_{ki} \left\vert \bs{\mu}_j, \bs{\Sigma}_j \right.
 \right),
\end{equation}
where $\mathbf{1}_{s_{ki} = s_j}$ is an indicator function which equals $1$ when $s_{ki} = s_j$ otherwise equals $0$. The indicator function draws from an intuitive principle: a measurement can only be assigned to a landmark with the same semantic label. This emphasizes semantic consistency in data association. Specifically, in implementation, the measurement $\bs{Z}$ is segmented into different layers based on semantic labels. Feature association and Gaussian fitting are then conducted within these designated layers separately, ensuring that each segmentation accurately contributes to the modeling of the environment (shown in Fig. \ref{fig_workflow}). The initial landmark states are obtained by voxelizing the initial map and treating each voxel as a Gaussian.

Considering the entire event space of $\bs{Y}$, the marginal likelihood of a measurement can be interpreted as a convex combination of $J$ components:
\begin{equation} \label{eq_gmm_rep}
    p\left( \hat{\bs{z}}_{ki} \left\vert \bs{T}, \bs{L}\right.  \right) = \sum_{j=1}^{J}\pi_j p\left( \hat{\bs{z}}_{ki} \left\vert \bs{T}, \bs{L}, y_{ki}=j \right. \right), 
\end{equation}
where $\sum_{j = 1}^J\pi_j=1$ is the mixing coefficient. 

\subsection{Expectation Maximization Registration}
\begin{figure}[!t]
\centering
\footnotesize
    \begin{subfigure}[t]{0.2\textwidth}
        \centering
        \includegraphics[width=\textwidth]{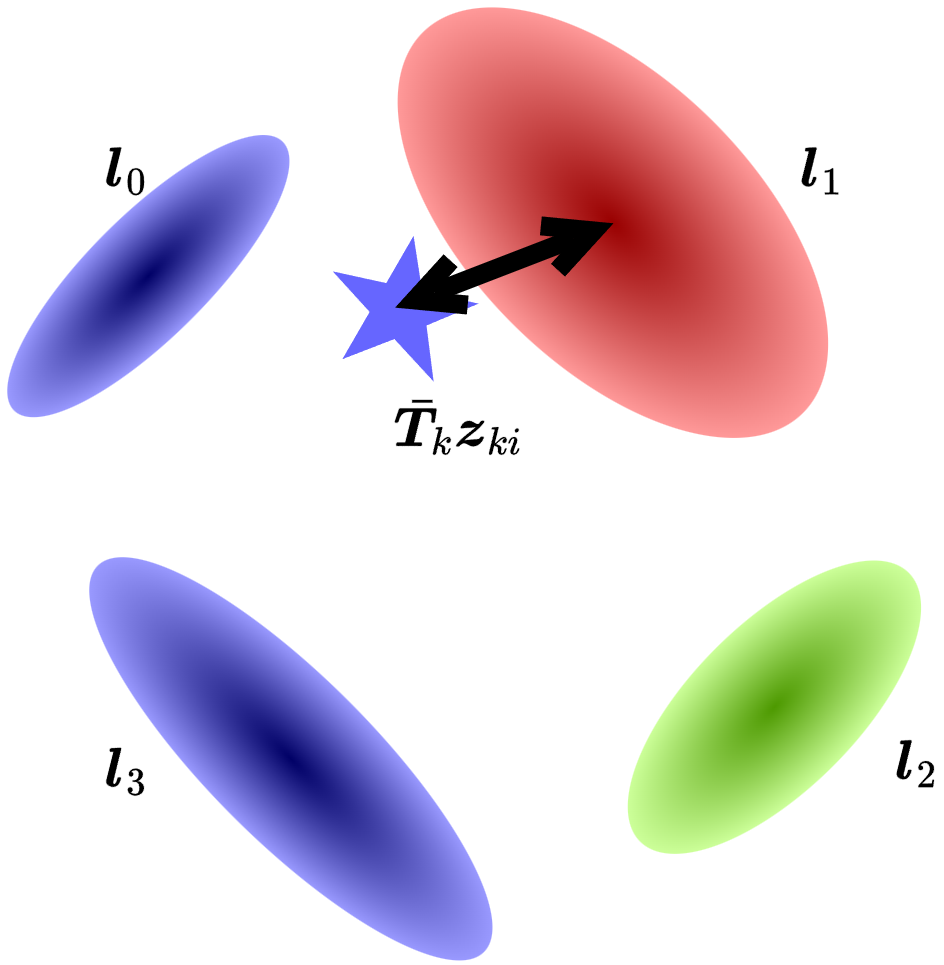}
        \caption{Deterministic association}
        \label{fig_comp_feature_asso_local}
    \end{subfigure}
    \quad 
    \begin{subfigure}[t]{0.2\textwidth}
        \centering
        \includegraphics[width=\textwidth]{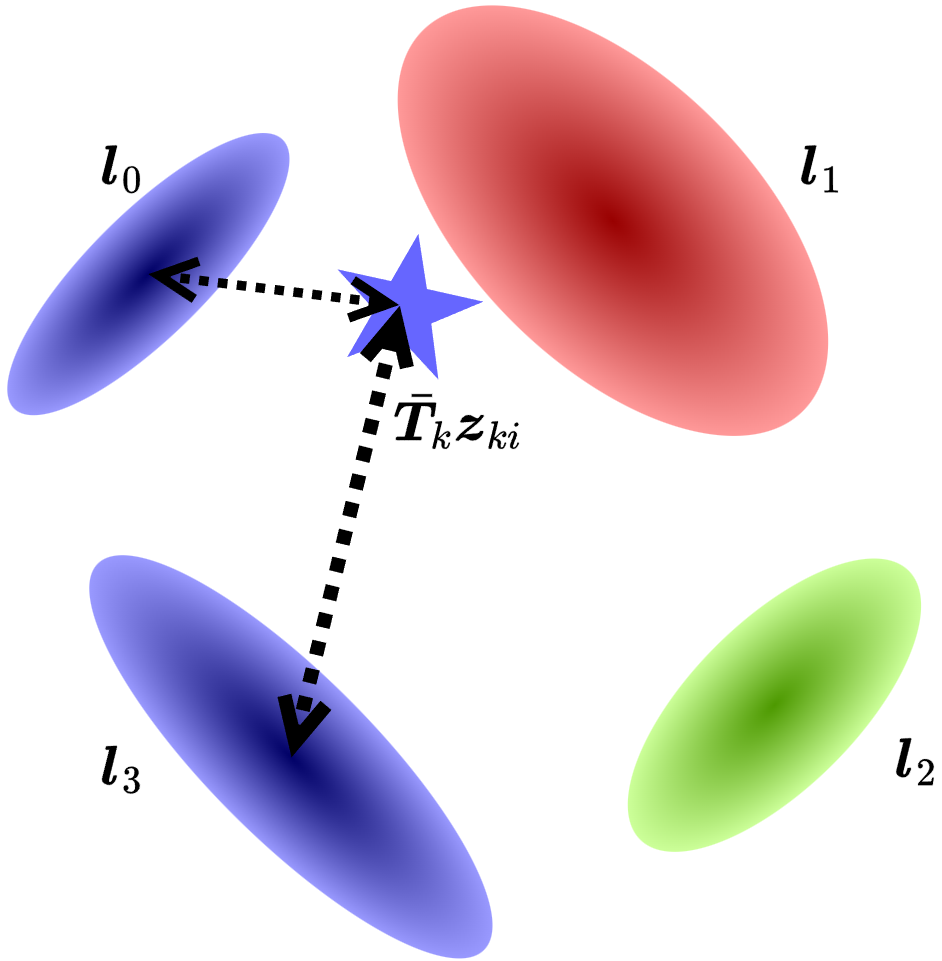}
        \caption{Probabilistic association}
        \label{fig_comp_feature_asso_global}
    \end{subfigure}
    \vspace{-5pt}
    \caption{Comparisons of feature association methods (different colors denote different semantic labels). (a). $\Bar{\bs{T}}_k\bs{z}_{ki}$ is directly assigned to the nearest but wrong landmark $\bs{l}_1$. (b). $\Bar{\bs{T}}_k\bs{z}_{ki}$ is soft-assigned to $\bs{l}_0$ and $\bs{l}_3$ with different probability.}
    \label{fig_local_global}
    \vspace{-15pt}
\end{figure}

To solve the MLE problem \eqref{eq_mle_ori}, existing works on LiDAR BA either assume that the optimal landmark states $\hat{\bs{L}}$ can be obtained as prior \cite{balm}, or adopt a ``coordinate descent'' approach which iterates the estimation of $\left\{\bs{T}, \bs{L}\right\}$ and $\bs{Y}$ separately \cite{balm2, ef, plc, multiview, pss_ba}. Regarding the estimation of feature association, the common approach in these works is maximum likelihood, where the most likely associations are computed and fixed during the iteration. However, this approach can be brittle, as incorrect associations can move the optimal solution far from the true values \cite{bowman2017probabilistic}. Furthermore, as illustrated in Fig. \ref{fig_local_global}, the deterministic association may lead the optimization to converge to local optima prematurely, thus preventing the attainment of global optima.

To tackle these problems, we propose to consider the entire probability density of $\bs{Y}$ when estimating $\bs{T}$ and $\bs{L}$, treating $\bs{Y}$ as a latent variable. That is, we choose to maximize the expected complete-data log-likelihood conditioned by the observed data. For convenience, the model parameter to be estimated will be denoted as $\bs{X} = \left\{ \left\{ \pi_j, \bs{\mu}_j, \bs{\Sigma}_j \right\}_{j=1}^J, \left\{ \bs{T}_k \right\}_{k=1}^K \right\}$. As a result, the MLE problem is reformulated as:
\begin{equation} \label{eq_mle_ecm}
\begin{split}
       \hat{\boldsymbol{T}}, \hat{\bs{L}} &= \operatorname*{argmax}_{\bs{T}, \bs{L}} \mathbb{E}_{\bs{Y}}\left( \log p\left( \bs{Z}, \bs{Y} ; \bs{T}, \bs{L} \right) \right) \\
    &= \operatorname*{argmax}_{\bs{T}, \bs{L}} \sum_{\bs{Y}} p\left( \bs{Y} \left \vert \bs{Z}, \bs{X} \right. \right) \log p\left( \bs{Z}, \bs{Y} ; \bs{X}\right). 
\end{split}
\end{equation}
Since $\bs{Y}$ is a latent variable, directly optimizing the MLE problem is impractical. Instead, we use the Expectation Maximization (EM) algorithm, which efficiently handles hidden variables and iteratively refines both the variables and model parameters, ensuring improved convergence and robustness.
\subsubsection{Expectation}
The first term of the objective function \eqref{eq_mle_ecm} represents the posterior probability of the associations given the measurements and model parameters. Denoted as $\alpha_{kij}$, it can be derived using Bayes' rule as follows:
\begin{equation} \label{eq_bayes}
\begin{split}
        \alpha_{kij} &= p\left( y_{ki}=j \left\vert \hat{\bs{z}}_{ki}, \bs{X} \right. \right) \\
        &=\frac{ p\left( \hat{\bs{z}}_{ki}, \bs{X} \left\vert y_{ki}=j \right.\right) p\left( y_{ki}=j \right)}{p\left( \hat{\bs{z}}_{ki}, \bs{X}\right)}.
\end{split}
\end{equation}
Recalling the GMM representation \eqref{eq_likelihood_single_measurement} and \eqref{eq_gmm_rep}, it is easy to have: 
\begin{equation} \label{eq_post}
    \alpha_{kij} = \frac{\mathbf{1}_{s_{ki} = s_j}\pi_j\mathcal{N}\left( \bs{T}_k\bs{z}_{ki} \left\vert \bs{\mu}_j, \bs{\Sigma}_j \right.
 \right)}{\sum_k \mathbf{1}_{s_{ki} = s_j} \pi_j \mathcal{N}\left( \bs{T}_k\bs{z}_{ki} \left\vert \bs{\mu}_j, \bs{\Sigma}_j \right.
 \right)}.
\end{equation}
In the expectation step, $\alpha_{kij}$ is calculated for all measurements. Notably, this calculation is performed exclusively within the specific semantic layer, as defined by the indicator function (see Fig. \ref{fig_comp_feature_asso_global}). In this way, we introduce the semantic constraints, significantly increasing computational efficiency and matching accuracy.

\subsubsection{Maximization}
The second term of the objective function \eqref{eq_mle_ecm} represents the complete-data log-likelihood. It can be expanded as follows:
\begin{equation} \label{eq_expand}
\begin{split}
    \log p\left( \bs{Z}, \bs{Y} ; \bs{X} \right)
    &= \log \prod_{k,i} p\left( \hat{\bs{z}}_{ki}, y_{ki} ; \bs{X} \right)p\left( y_{ki} \right) \\
    &= \log \prod_{k,i,j} \left( \pi_{j}p\left( \hat{\bs{z}}_{ki} \left\vert y_{ki}=j, \bs{X}\right. \right) \right)^{\mathbf{1}_{s_{ki} = s_j}}.
\end{split}
\end{equation}
Subjecting \eqref{eq_post} and \eqref{eq_expand} back to \eqref{eq_mle_ecm}, we have: 
\begin{equation}
    \begin{split}
        \hat{\boldsymbol{T}}, \hat{\bs{L}} &= \operatorname*{argmax}_{\bs{T}, \bs{L}} \sum_{k, i, j}\alpha_{kij}\left( \log\pi_j + \log p\left(\bs{z}_{ki}\left\vert \bs{X}, y_{ki}=j \right.  \right) \right) \\
        &= \operatorname*{argmin}_{\bs{T}, \bs{L}}\sum_{k, i, j}\alpha_{kij}\left( \| \bs{T}_k\bs{z}_{ki} - \bs{\mu}_j \|_{\bs{\Sigma}_j}^{2} + \log\lvert \bs{\Sigma}_j \rvert\right).
    \end{split}
\end{equation}
The landmark states $\bs{\mu}_j$ and $\bs{\Sigma}_j$ are functions of the transformation matrices $\bs{T}_k \in \mathrm{SE}(3)$. As a result, simultaneously estimating all elements in $\bs{X}$ results in a challenging nonlinear minimization problem. Instead, we propose estimating $\bs{T}$ and $\bs{L}$ separately, keeping one constant while optimizing the other. This approach replaces the EM algorithm with Expectation Conditional Maximization (ECM) \cite{rigid_articulated}. Specifically, we first estimate $\bs{T}$ by:
\begin{equation} \label{eq_maxmization_t}
        \hat{\bs{T}} = \operatorname*{argmin}_{\bs{T}}\sum_{k, i, j}\alpha_{kij} \left\| \bs{T}_k\bs{z}_{ki} - \Bar{\bs{\mu}}_j \right\|_{\Bar{\bs{\Sigma}}_j}^{2},
\end{equation}
where $\Bar{\bs{\mu}}_j, \Bar{\bs{\Sigma}}_j$ are the prior estimates. This step aims to estimate the optimal transformation matrices that minimize the Mahalanobis distances between points $\bs{z}_{ki}$ and the distributions $\left(\Bar{\bs{\mu}}_j, \Bar{\bs{\Sigma}}_j\right)$ weighted by the posterior probabilities $\alpha_{kij}$. To further improve the efficiency, a virtual measurement, denoted as $\bs{w}_{ki}$ is introduced:
\begin{equation} \label{eq_}
        \bs{w}_{ki} = \frac{\sum_{i} \alpha_{kij}\bs{z}_{ki}}{\sum_{i} \alpha_{kij}}, 
\end{equation}
which represents the weighted average of all the points in LiDAR scan $\bs{Z}_k$ assigned to the landmark $\bs{l}_j$. As a result, \eqref{eq_maxmization_t} can be reduced to:
\begin{equation} \label{eq_maxmization_t_reduced}
        \hat{\bs{T}} = \operatorname*{argmin}_{\bs{T}}\sum_{k, j}\beta_{kj} \left\| \bs{T}_k\bs{w}_{kj} - \Bar{\bs{\mu}}_j \right\|_{\Bar{\bs{\Sigma}}_j}^{2},
\end{equation}
where $\beta_{kj}=\sum_i\alpha_{kij}$. This forms a standard least squares problem and can be solved using the method proposed in VGICP \cite{vgicp}. 

After having the optimial estimate $\hat{\bs{T}}$, the landmark states $\hat{\bs{L}}$ can be updated as \cite{rigid_articulated}:
\begin{equation} \label{eq_m_m}
\begin{split}
        \hat{\bs{\mu}}_j &= \sum_{k,i}\frac{\hat{\pi}_j\hat{\bs{T}}_k\bs{z}_{ki}}{\hat{\pi}_j}, \\
        \hat{\bs{\Sigma}}_j &= \sum_{k, i}\frac{\hat{\pi}_j \alpha_{kij} \left( \hat{\bs{T}}_k\bs{z}_{ki} - \hat{\bs{\mu}}_j \right) \left( \hat{\bs{T}}_k\bs{z}_{ki} - \hat{\bs{\mu}}_j \right)^T}{\hat{\pi}_j}.
\end{split}
\end{equation}
Specifically, the mixture parameter $\pi_j$ cannot be updated following the standard method as the summation of the posterior probability. This approach would result in $\pi_j$ dominated by the point amount, and thus the semantic layer with more points will dominate the optimization process. To ensure equal contribution from all semantic layers, the mixture parameter is fixed as $\pi_j= 1 / N_s J_s$, where $N_s$ is the number of semantic layers and $J_s$ denotes the number of Gaussian mixtures within the semantic layer $s_j$. 

\subsection{Adaptive Semantic Label Selection} \label{sec_label_select}

Selecting all semantic layers to construct the cost function is unnecessary and consumes high computation resources. On the other hand, solely adopting specific semantic layers results in limited generalizability, similar to the issues with predefined geometric features. Thus, we wish to find an approach that utilizes the fewest labels possible while ensuring the optimization is well-conditioned in all directions. This requires a metric that can measure the degree of degeneracy. To solve the MLE problem \eqref{eq_maxmization_t_reduced} we linearize it at the prior estimate $\Bar{\bs{T}}$ to have the error state $\delta\bs{T}$:
\begin{equation}
    \delta\bs{T}=\begin{bmatrix}
    \dots && \delta\bs{r}_k^T && \delta\bs{p}_k^T && \dots
    \end{bmatrix}^T \in \mathbb{R}^{6K},
\end{equation}
where $\delta\bs{r}_k = Log\left( \hat{\bs{R}}_k\Bar{\bs{R}}_k^{-1} \right) \in \mathbb{R}^3$ and $\delta\bs{p}_k =\hat{\bs{p}}_k-\Bar{\bs{p}}_k$. The MLE problem thus becomes a linear least squares problem:
\begin{equation} \label{eq_linear}
    \delta\bs{T} = \operatorname*{argmin}_{\delta\bs{T}}\left\|\bs{H}\delta\bs{T}-\bs{b}  \right\|^2, 
\end{equation}
This problem is about solving for $\delta\bs{T}$ based on the residual $\bs{b}$ and Jacobian matrix $\bs{H}$. We study the condition number of the problem with respect to perturbations in $\bs{H}$ \cite{bookLinearAlgebra}: 
\vspace{-3pt}
\begin{equation}
    \kappa\left(\bs{H}\right) = \frac{\sigma_{max}\left\{\bs{H}\right\}}{\sigma_{min}\left\{\bs{H}\right\}} = \sqrt{\frac{\lambda_{max}\left\{\bs{H}^T\bs{H}\right\}}{\lambda_{min}\left\{\bs{H}^T\bs{H}\right\}}},
\end{equation}
where $\lambda_{max}\left\{\bs{H}^T\bs{H}\right\}$ and $\lambda_{min}\left\{\bs{H}^T\bs{H}\right\}$ represent the maximum and minimum eigenvalue of $\bs{H}^T\bs{H}$. $\kappa\left(\bs{H}\right)$ is important in numerical linear algebra, for it can determine how accurately one can solve the linear least squares problem. If a problem $\bs{H}\delta\bs{T}=\bs{b}$ has a high $\kappa\left(\bs{H}\right)$, it is ill-conditioned and should expect to ``lose $\log\kappa\left(\bs{H}\right) digits$'' in computing the solution \cite{bookLinearAlgebra}. Therefore, our goal can be summarized as follows: we want to construct \eqref{eq_maxmization_t_reduced} with as few constraints as possible while ensuring that $\kappa\left\{\bs{H}\right\}$ is not too large (usually less than 100 is considered well-conditioned \cite{golub13}). Compared with other degeneracy evaluation frameworks \cite{zhang2016degeneracy, tuna2023x}, conditional number eliminates the need for scenario-specific thresholds and provides computational efficiency, reducing the overhead associated with methods like information analysis.

In implementation, we divide the entire semantic set $\bs{S}$ into two parts: the initial set $\bs{S}_{ini}$ and the remaining set $\bs{S}_{rm}$. $\bs{S}_{ini}$ should include ubiquitous semantic layers in the operating environment that ensure a well-conditioned optimization problem, such as ground, poles, and lane markings for autonomous vehicles. The optimization problem \eqref{eq_maxmization_t_reduced} is initially constructed on $\bs{S}_{ini}$. Then we linearize it at $\Bar{\bs{T}}$ and evaluate the condition number $\kappa\left\{\bs{H}\right\}$. If $\kappa\left\{\bs{H}\right\} < \kappa_{thd}$, the problem is well-conditioned on $\bs{S}_{ini}$. Otherwise, the problem is considered degenerate and we need to randomly select a semantic label $s$ from $\bs{S}_{rm}$ and combine it with $\bs{S}_{ini}$ to construct a new cost function. $s$ will be retained if the new condition number is smaller than the original one; otherwise, it will be discarded. This process is repeated until the new condition number falls below $\kappa_{thd}$. If this cannot be achieved even after $n_c$ iterations, the poses will remain unchanged. In conclusion, the implementation of SGBA on a batch of LiDAR scans is illustrated in Algorithm \ref{alg_GBA}. To clarify, $\bs{X}^q = \left\{\bs{L}^q, \bs{T}^q \right\}$ denotes the model parameter at the $q_{th}$ iteration.
\begin{algorithm}
    \caption{SGBA for a batch of LiDAR scans}\label{alg_GBA}
\KwData{$\bs{Z}$, $v$, $\Bar{\bs{T}}$, $\left\{ \pi_j \right\}_{j=1}^J$, $\bs{S}_{ini}$, $\kappa_{thd}$, $n_c$}
\KwResult{$\hat{\bs{T}}, \hat{\bs{L}}$}
\tcp{\scriptsize Voxelize each semantic layer separately}
$\left\{\bs{\mu}^0_j, \bs{\Sigma}^0_j\right\} \gets \text{voxelization}(\bs{Z},\Bar{\bs{T}}, v)$\;
$q \gets 1$, $n \gets 0$\;  
\tcp{\scriptsize E-step}
Update $\left\{\alpha_{jik}\right\}$ by $\bs{X}^{q-1}$ and $\bs{S}_{ini}$, i.e. \eqref{eq_post}\;
Construct \eqref{eq_maxmization_t_reduced} with $\left\{\alpha_{jik}\right\}$, $\bs{X}^{q-1}$ and $\bs{S}_{ini}$ \;
Linearize \eqref{eq_maxmization_t_reduced} at $\Bar{\bs{T}}$ to have $\bs{H}$\;

\While{$not$ $converged$}{
    \While{$n < n_c$}{
        $n \gets n + 1$\;
        \If{$\kappa\left\{\bs{H}\right\} < \kappa_{thd}$}{
            \textbf{break}\;
        } \Else {
            Calculate new Jacobian: $\bs{H}'$\ with $\bs{S}_{ini} \cup\{s\}, s \in \bs{S}_{rm}$\;
            \If{$\kappa\left\{\bs{H}'\right\} < \kappa\left\{\bs{H}\right\}$}{
                $\bs{S}_{ini} \gets \bs{S}_{ini} \cup \{s\}$\;
                $\kappa\left\{\bs{H}\right\} \gets \kappa_{new}\left\{\bs{H}\right\}$\;
            }
        }
    }

    \If{$\kappa\left\{\bs{H}\right\} > \kappa_{thd}$}{
        \textbf{return} $\Bar{\bs{T}}, \Bar{\bs{L}}$\;
    }
    
    \tcp{\scriptsize M-step}
    Estimate $\hat{\bs{T}}^q$ by $\alpha_{jik}$, i.e. \eqref{eq_maxmization_t_reduced}\;
    Estimate $\hat{\bs{L}}^q$ by $\hat{\bs{T}}^q$, i.e. \eqref{eq_m_m}\;
    $q \gets  q + 1$\;
}
\end{algorithm}

\section{Experiments}
We implement the proposed SGBA into a system that performs local BA on a sliding window, as illustrated in Fig. \ref{fig_workflow}. The input is the prior poses $\Bar{\bs{T}}$ estimated by LiDAR odometry, the corresponding scans $\bs{Z}$, and the semantic labels $\bs{S}$. We set the window size $w$ to 10 and only the keyframes are accepted for the BA process. The initial set $\bs{S}_{ini}$ is : $\bs{S}_{ini}=\left\{ car, road, pole, lane\text{-}marking, trunk \right\}$. The condition number threshold $\kappa_{thd}$ and iteration threshold $n_c$ are set correspondingly to 100 and 6. The voxel size $v$ for initialization is selected as 6$m$ for $ground$ and 3$m$ for other labels. To evaluate the robustness, accuracy, and efficiency of SGBA, we have conducted various experiments on different datasets. All experiments were conducted using the same parameters on a laptop with a 2.3 GHz AMD Ryzen 7 6800H CPU and 16 GB RAM. 
\subsection{Kitti Dataset}

\begin{figure}[!t]
\centering
\includegraphics[width=3.4in]{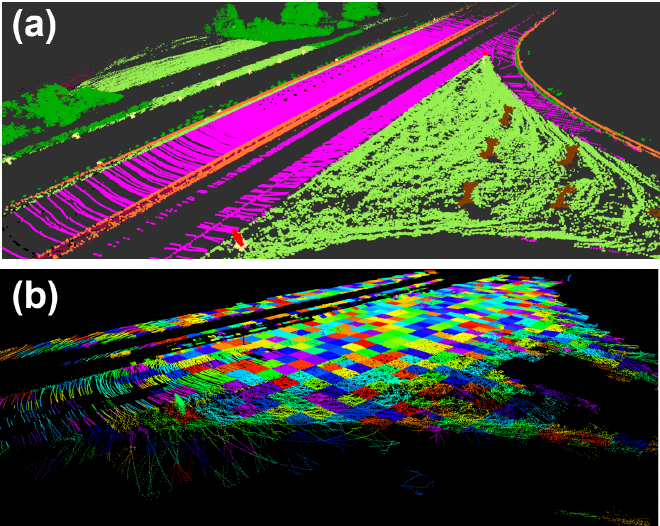}
\vspace{-5pt}
\caption{Scene dominated by degenerative planar features causing geometric-only LiDAR BA to fail. (a). Original point cloud. (b). Planes extracted by plane-based BA methods. No planes can provide horizontal constraints.}
\label{fig_01_no_horizon}
\vspace{-10pt}
\end{figure}

The Kitti dataset is collected with a Velodyne-64 LiDAR mounted on a vehicle, including residential and highway environments \cite{kitti}. We compare SGBA with a SOTA method, BALM \cite{balm2}, a LiDAR BA method extracting planar as landmarks. We modified the open-sourced code of BALM so that it can perform local BA on a sliding window. The prior estimate $\Bar{\bs{T}}$ is obtained by a LiDAR odometry method, kiss-ICP \cite{vizzo2023kiss}. The semantic labels are predicted using SPVNAS \cite{spvnas}, a lightweight 3D neural network achieving around $60\%$ mIoU for outdoor scene segmentation in near real-time.  To evaluate the influence of the semantic labels on the performance, we feed SGBA with two kinds of semantic labels: SGBA($100\%$) takes the ground truth labels as input, while SGBA($60\%$) takes the labels predicted by SPVNAS. The RMSE of the absolute trajectory error (ATE) \cite{grupp2017evo} is used to evaluate the accuracy of the estimated poses.

As illustrated in Tab. \ref{table:kitti}, SGBA(gt) achieves the best accuracy and SGBA achieves the second-best results with an acceptable difference of only 0.07 meters. When evaluating each sequence individually, we find that SGBA(gt) does not consistently outperform SGBA. Therefore, we conclude that the accuracy of semantic segmentation contributes to the overall accuracy of SGBA, but its influence is subtle if the mIoU is greater than $60\%$.

Compared to BALM, SGBA demonstrates better accuracy for most sequences. This is particularly evident in sequence 01. The reason is that BALM only extracts planar features as landmarks, while sequence 01 was collected in a highway environment where few planes provide horizontal constraints (see Fig. \ref{fig_01_no_horizon}). As a result, BALM degenerates in the horizontal direction and is more susceptible to noise \cite{zhang2016degeneracy}. Instead, SGBA models the environment as a semantic GMM without relying on predefined feature types and thus can utilize the poles and trunks for horizontal constraints. For other sequences, BALM and SGBA produce comparable results. This is because, in these environments, the presence of many planar features provides sufficient constraints for BALM, thus minimizing the differences between the two methods. In conclusion, SGBA achieves more accurate and robust performance than BALM. With the semantic GMM, SGBA can effectively handle a variety of environments, achieving accurate and reliable pose estimation. 

\begin{table}[!t]
\centering
\caption{The RMSE of ATE (Meters) for Kitti dataset}
\label{table:kitti}
\setlength{\tabcolsep}{.7pt} 
\begin{threeparttable}
\begin{tabularx}{\linewidth}{>{\raggedright\arraybackslash}X*{5}{>{\centering\arraybackslash}X}}
\toprule
         & 01 & 05    & 06 & 07 & Avg.\\ 
\midrule                                                                                                                          
Kiss-ICP & 9.91 & 2.19 & 1.21 & 1.10 & 3.34\\
BALM & 10.69 & \underline{1.81} & 1.24 & \underline{0.92} & 3.67\\
SGBA($100\%$) & \textbf{8.74}  & \textbf{1.69} & \textbf{1.06}  & {0.94}& \textbf{3.11}\\
SGBA($60\%$) & \underline{8.95}  & \textbf{1.69} & \underline{1.13}  & \textbf{0.88}& \underline{3.16}\\
\bottomrule
\end{tabularx}
\begin{tablenotes}
    \footnotesize
        \item Avg. denotes the average ATEs for all the sequences.
    \item \textbf{Bold}: the best result; \underline{underline}: second best.
\end{tablenotes}
\end{threeparttable}
\vspace{-15pt}
\end{table}

\subsection{MCD Dataset}

\begin{figure*}[!t]
\centering
\includegraphics[width=6.8in]{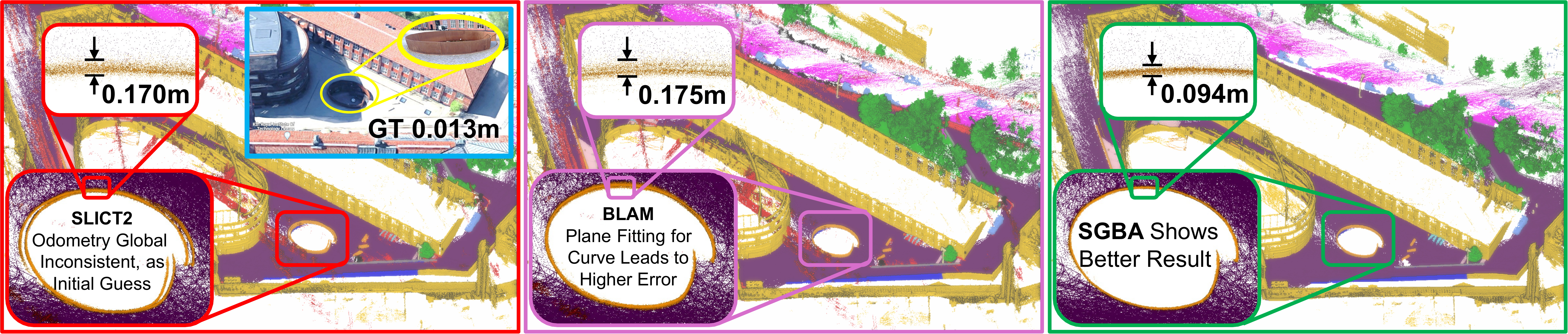}
\caption{Left: SLICT, Middle: BALM, Right: SGBA. Map comparison on kth\_05 sequence using MCD semantic coloring.  SGBA shows better generalization for the challenging outdoor curved scene. }
\label{fig_map_compare}
\vspace{-10pt}
\end{figure*}

The MCD dataset was collected in multiple campuses with two LiDARs: an os-128 and a Livox mid-70. The ntu\_* sequences are collected on an electric vehicle and the kth\_* and tuhh\_* sequences are collected with a hand-held device. The prior estimate $\Bar{\bs{T}}$ is obtained by SLICT \cite{nguyen2023SLICT}, a LiDAR-Inertial odometry that models the environment as surfels. For os-128, we use SPVNAS trained on the Kitti dataset to predict the semantic labels, with mIoU less than $60\%$. For Livox mid-70, we directly use the ground truth semantic labels.

\subsubsection{OS-128 LiDAR} \label{sec_os128}
In this section, we evaluate the influence of the probabilistic feature association strategy on the accuracy of pose estimation. We change the formulation of the posterior probability $\alpha_{kij}$ such that the point $\bs{z}_{ki}$ is assigned directly to its nearest landmark $\bs{l}_j$ with the same semantic label. The resulting system is denoted as SGBA(near). Moreover, we assess the impact of the prior estimate pose quality by using different IMUs. Sequences utilizing the built-in IMU of the os-128 are denoted as ntuo\_*, while those using the V100 IMU are denoted as ntuv\_*. 

As illustrated in Tab. \ref{table:os128}, the average accuracy of SGBA outperforms that of SGBA(near), which indicates that the probabilistic association method can better handle uncertainties and ambiguities in the LiDAR data and prior poses. This probabilistic strategy ensures that the optimization will not converge too early at the local optimum. Both SGBA and SGBA(near) outperform BALM in terms of average accuracy, demonstrating the effectiveness of incorporating semantic constraints in improving accuracy. Specifically, the difference in accuracy between SGBA and BALM is particularly obvious on ntuo\_* sequences at 0.47m, compared to 0.03m on ntuv\_* sequence. This is because BALM relies on good prior pose estimates to initialize the correct plane features. Otherwise, the false features will move the optimal solution from the true value. In contrast, SGBA relies primarily on semantic labels for clustering, enabling it to better handle suboptimal initial poses and maintain higher accuracy in pose estimation. In conclusion, with the probabilistic association and semantic GMM representation, SGBA demonstrates robustness and accuracy even in challenging scenarios with lower-quality initial pose estimates.

\begin{table}[t]
\centering
\caption{The RMSE of ATE (Meters) for os-128}
\label{table:os128}
\setlength{\tabcolsep}{1pt} 
\begin{threeparttable}
\begin{tabularx}{\linewidth}{>{\raggedright\arraybackslash}X*{5}{>{\centering\arraybackslash}X}}
\toprule
         & ntuo\_{01} & ntuo\_{13}    & ntuv\_{01} & ntuv\_{13} & Avg.\\ 
\midrule                                                                                                                          
SLICT & 4.77 & \underline{2.39} & 2.19 & 1.15 & 2.62\\
BALM & 4.63 & 2.52 & 1.96 & \textbf{1.04} & 2.48\\
SGBA(near) & \underline{4.41}  & {2.43} & \underline{1.88} & \underline{1.10}& \underline{2.45}\\
SGBA & \textbf{4.29}  & \textbf{1.93} & \textbf{1.80} & {1.14}& \textbf{2.28}\\
\bottomrule
\end{tabularx}
\begin{tablenotes}
    \footnotesize
    \item Avg. denotes the average ATEs for all the sequences.
    \item \textbf{Bold}: the best result; \underline{underline}: second best.
\end{tablenotes}
\end{threeparttable}
\vspace{-10pt}
\end{table}

\subsubsection{Livox mid-70 LiDAR} \label{sec_livox}

To evaluate the influence of the proposed adaptive semantic selection scheme, we modify SGBA to have 2 variants: SGBA(min) utilizes only the initial semantic set $\bs{S}_{ini}$ for optimization, corresponding to the strategy of solely adopting predefined semantic labels; conversely, SGBA(max) employs all available semantic labels. The results are shown in Tab. \ref{table:mid70}, it is obvious that all the variants get worse compared to the original and SGBA outperforms BALM on these sequences (see Fig. \ref{fig_map_compare}). For SGBA(max), the results either diverge or deteriorate compared to the prior estimates. On the one hand, the inclusion of unstable semantic labels such as pedestrians, moving cars, and vegetation introduces instability into the optimization process. On the other hand, adopting all semantic labels introduces too many redundant constraints and thus makes the problem computationally infeasible to solve. SGBA(min) achieves similar performance to SGBA on ntu\_* sequences but diverges on kth\_* and tuhh\_* sequences. This discrepancy arises from the fact that $\bs{S}_{ini}$ comprises only the essential semantic clusters for autonomous vehicle environments. While ntu\_* sequences are captured by an autonomous vehicle, kth\_* and tuhh\_* sequences are collected using hand-held devices, existing scenarios where the labels in $\bs{S}_{ini}$ may be totally absent. Consequently, for kth\_* and tuhh\_* sequences, the initial set lacks enough constraints to formulate a well-conditioned optimization problem and leads to divergence.

In conclusion, the adaptive semantic selection scheme in SGBA proves to be essential for achieving optimal performance. Dynamically selecting the most informative and stable semantic clusters ensures that the optimization process remains well-conditioned, thereby enhancing the robustness and accuracy of pose estimation in diverse environments.

\subsection{Computational Efficiency} 
In this section, we evaluate the computational efficiency of SGBA. To further assess the impact of the adaptive semantic selection scheme, we define two variants of SGBA as described in Section \ref{sec_livox}.

\begin{figure}[!t]
    \centering
    \subfloat[Evaluating on Livox Mid-70]{%
        \includegraphics[width=0.40\textwidth]{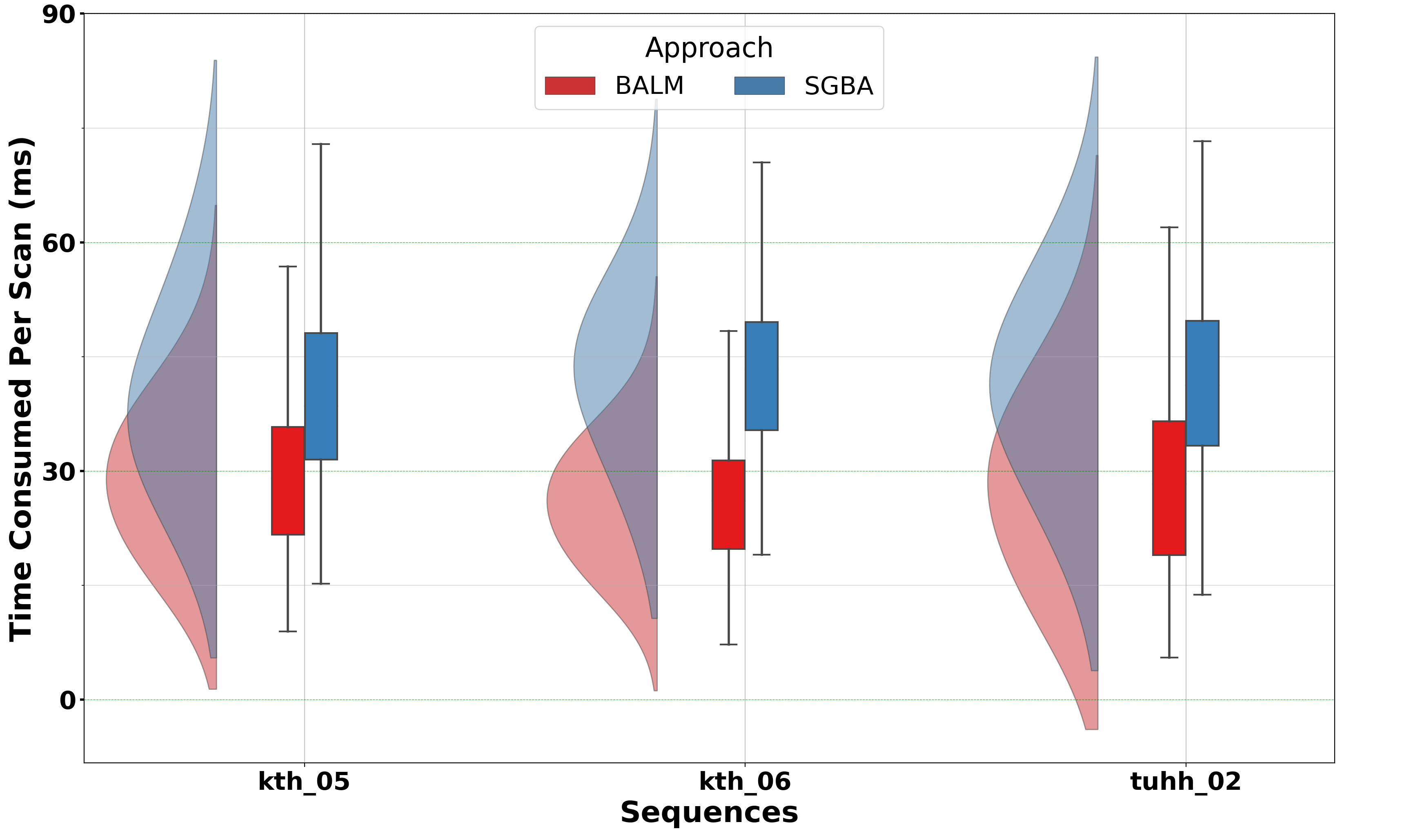}}
    \hfill
    \subfloat[Evaluating on Velodyne-64]{%
        \includegraphics[width=0.40\textwidth]{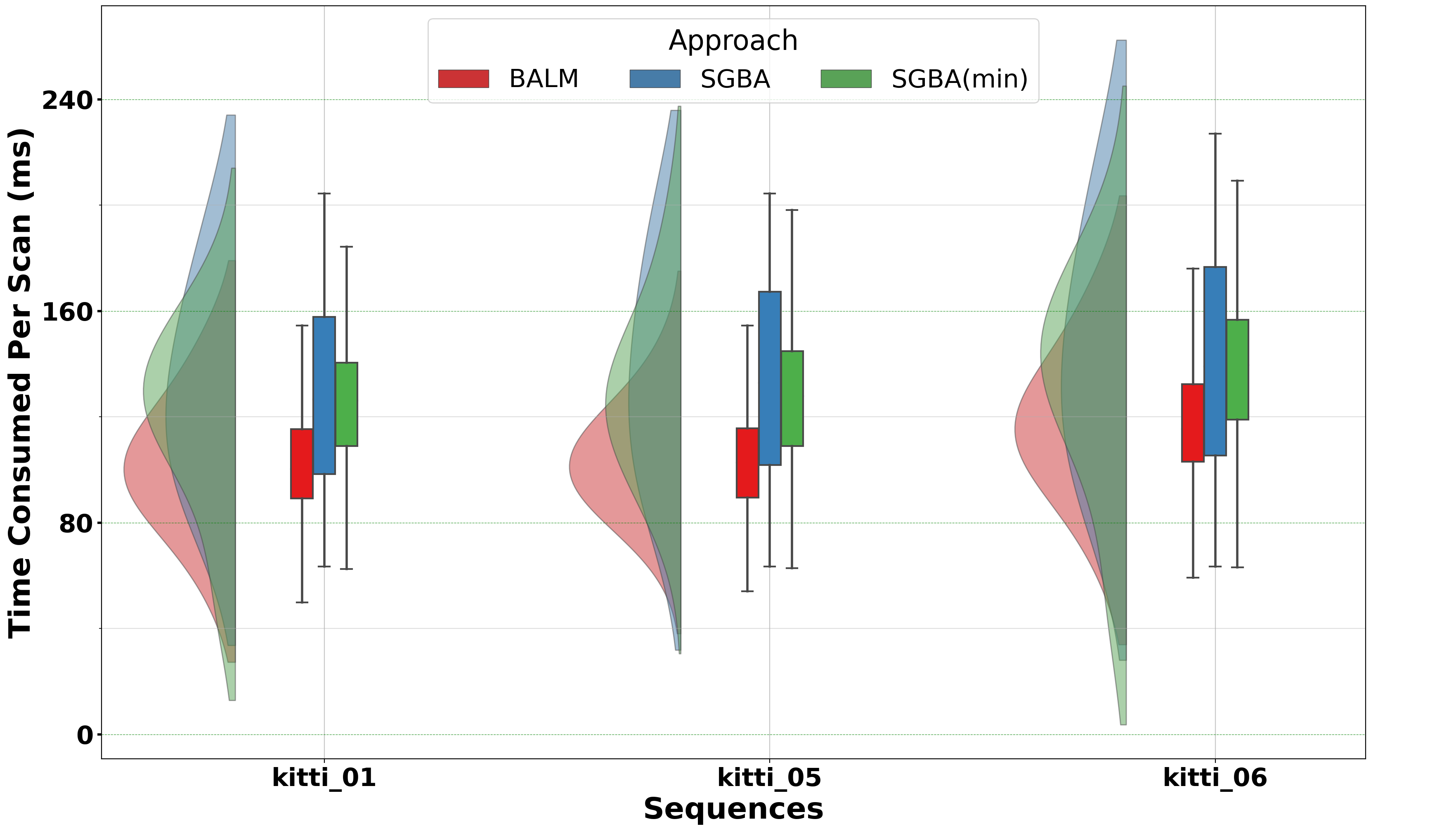}}
    \caption{Comparison of average scan times for different methods shows that adding semantics increases computational cost but remains comparable to BALM.}
    \label{fig_time}
    \vspace{-5pt}
\end{figure}

The results are listed in Fig. \ref{fig_time}. The time consumed per scan is calculated by dividing the total processing time of the sliding window by the number of frames within. It should be noted that the average processing time for SGBA(max) is significantly greater than 250ms thus it is not visualized to avoid skewing the figure. SGBA(max) consumes the highest time due to the inclusion of too many redundant constraints, which significantly increases the dimension of the cost function and thus drastically increases the time complexity needed to solve it. SGBA takes slightly longer than SGBA(min) for sequences where SGBA(min) does not diverge. This is because solving \eqref{eq_maxmization_t_reduced} requires the calculation of $\bs{H}^T\bs{H}$, and thus the additional time is primarily spent performing Singular Value Decomposition (SVD) on $\bs{H}^T\bs{H}$, which does not consume excessive time since the dimension is limited. Therefore, the marginal increase of time required to evaluate the degeneracy is acceptable, as it can prevent divergence.

BALM consumes the shortest time for all sequences due to the efficiency of planar model fitting. Additionally, the introduction of ``point clusters'' further enhances the optimization process \cite{balm2}. In conclusion, the adaptive semantic selection scheme in SGBA provides a balance between computational speed and robustness, ensuring accurate pose estimation while minimizing the risk of divergence. Although BALM demonstrates superior speed, SGBA's performance remains acceptable, and it achieves better robustness and accuracy for most sequences.

\begin{table}[!t]
\centering
\caption{The RMSE of ATE (Meters) for mid-70}
\label{table:mid70}
\setlength{\tabcolsep}{1pt} 
\begin{threeparttable}
\begin{tabularx}{\linewidth}{>{\raggedright\arraybackslash}X*{5}{>{\centering\arraybackslash}X}}
\toprule
         & ntu\_{01} & ntu\_{13}  & kth\_{05} & kth\_{06} & tuhh\_02\\ 
\midrule                                                                                                                          
SLICT       & 4.38 & \underline{1.74} & 1.63 & 1.33 & \underline{3.16}\\
BALM & \underline{4.19} & {1.91} & \underline{1.52} & \underline{1.18} & 3.19\\
SGBA(max) & $\times$  & {2.00} & 1.85 & {1.49}& $\times$\\
SGBA(min) & \textbf{3.95}  & \textbf{1.39} & $\times$ & $\times$& $\times$\\
SGBA & \textbf{3.95}  & \textbf{1.39} & \textbf{1.22} & \textbf{1.04}& \textbf{2.73}\\
\bottomrule
\end{tabularx}
\begin{tablenotes}
    \footnotesize
    \item \textbf{Bold}: the best result; \underline{underline}: second best.
    \item $\times$ denotes that the system diverged.
\end{tablenotes}
\end{threeparttable}
\vspace{-15pt}
\end{table}

\section{Limitations and Future Works}
Despite the improved accuracy and robustness of SGBA, there exist drawbacks that need further enhancement. One major challenge is \textbf{efficiency}, particularly regarding the computational resources required for calculating the posterior probability. One potential solution is constructing a more efficient data structure, such as ikd-tree \cite{xu2022fast} to accelerate the calculation. 

Another drawback is that the current approach is only a local BA method and cannot guarantee global consistency. Future work could focus on utilizing semantic Gaussians for loop detection to build a globally consistent system.

With the introduction of open vocabulary segmentation approaches, there is great potential for SGBA ahead where unsupervised semantic label and instances can be to be used for BA. We will explore the methods, such as Segment Anything Model 3D \cite{yang2023sam3d}, to test open vocabulary segmentation-based SGBM in the near future.

\section{Conclusion}
In this paper, we present a novel LiDAR BA approach that integrates both geometric information and semantic attributes to enhance pose estimation for autonomous systems. By modeling the environment as a semantic GMM, our method provides a flexible and general representation that does not rely on predefined features, making it adaptable to various environments. Secondly, we propose an adaptive semantic selection framework that evaluates the condition number of the optimization problem to determine the most informative semantic layers to use during the BA process. This framework helps prevent optimization degeneracy while using the fewest constraints possible, balancing generalizability and efficiency. Lastly, we address the challenges of accurate feature association by introducing a probabilistic approach, which considers the entire probability density of feature assignments and introduces semantic constraints. This approach allows for better handling of uncertainties and ambiguities in LiDAR data and initial pose estimates. Our experimental results demonstrate that the proposed SGBA approach SOTA methods in terms of accuracy and robustness across different datasets.

\bibliographystyle{IEEEtran}
\bibliography{mybib}

\end{document}